\documentclass[final]{cvpr}
\usepackage{times}
\usepackage{epsfig}
\usepackage{graphicx}
\usepackage{amsmath}
\usepackage{amssymb}
\usepackage{subfig}
\usepackage{bm} 
\usepackage{multirow}
\usepackage{makecell}
\usepackage{url}
\usepackage[pagebackref=true,breaklinks=true,colorlinks,bookmarks=false]{hyperref}

\def\cvprPaperID{8088} % *** Enter the CVPR Paper ID here
\def\confYear{CVPR 2021}

\begin{document}

%%%%%%%%% TITLE
\title{DCT-Mask: Discrete Cosine Transform Mask Representation for \\ Instance Segmentation}

\author{Xing Shen\textsuperscript{1,2}\footnotemark[1], Jirui Yang\textsuperscript{2}\footnotemark[1], Chunbo Wei\textsuperscript{2}, Bing Deng\textsuperscript{2}, Jianqiang Huang\textsuperscript{2}, Xiansheng Hua\textsuperscript{2}\\
Xiaoliang Cheng\textsuperscript{1}, Kewei Liang \textsuperscript{1}\footnotemark[2]\\
\textsuperscript{1}School of Mathematical Sciences, Zhejiang University \space \textsuperscript{2}DAMO Academy, Alibaba Group\\
{\tt\small \textsuperscript{1}\{shenxingsx,xiaoliangcheng,matlkw\}@zju.edu.cn}\\
{\tt\small \textsuperscript{2} \{jirui.yjr,chunbo.wcb,dengbing.db,jianqiang.hjq,xiansheng.hxs\}@alibaba-inc.com}\\
}
\maketitle

\renewcommand{\thefootnote}{\fnsymbol{footnote}}
\footnotetext[1]{These authors contributed equally to this work.} 
\footnotetext[2]{Corresponding author.}

\pagestyle{empty}  % no page number for the second and the later pages
\thispagestyle{empty} % no page number for the first page

%%%%%%%%% ABSTRACT
\begin{abstract}
%突出提出问题的思路
Binary grid mask representation is broadly used in instance segmentation. A representative instantiation is Mask R-CNN which predicts masks on a $28\times 28$ binary grid. Generally, a low-resolution grid is not sufficient to capture the details, while a high-resolution grid dramatically increases the training complexity. In this paper, we propose a new mask representation by applying the discrete cosine transform(DCT) to encode the high-resolution binary grid mask into a compact vector. Our method, termed DCT-Mask, could be easily integrated into most pixel-based instance segmentation methods. Without any bells and whistles, DCT-Mask yields significant gains on different frameworks, backbones, datasets, and training schedules. It does not require any pre-processing or pre-training,  and almost no harm to the running speed. Especially, for higher-quality annotations and more complex backbones, our method has a greater improvement. Moreover, we analyze the performance of our method from the perspective of the quality of mask representation. The main reason why DCT-Mask works well is that it obtains a high-quality mask representation with low complexity.
Code is available at \url{https://github.com/aliyun/DCT-Mask.git}
\end{abstract}

\section{Introduction}
\begin{figure}
\centering
\captionsetup[subfloat]{labelformat=empty}
\scalebox{1.0}{
\subfloat[(a) Binary grid mask representation]{
\begin{minipage}[t]{0.9\linewidth}
\captionsetup[subfloat]{labelformat=empty}
\centering
\subfloat[Ground Truth]{
\includegraphics[width=0.24\linewidth]{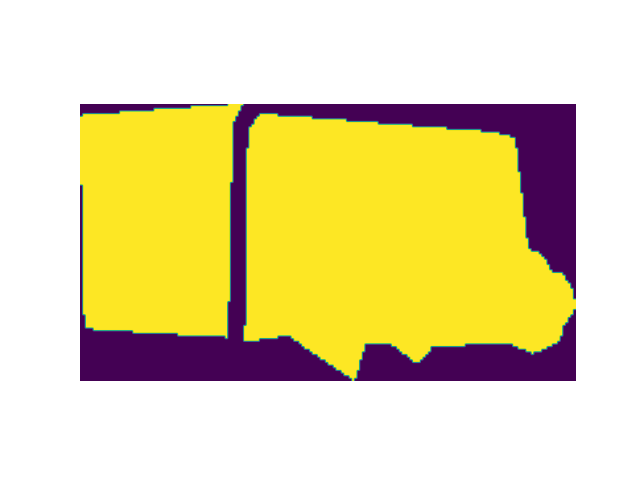}
}
\subfloat[$28\times 28$]{
\includegraphics[width=0.24\linewidth]{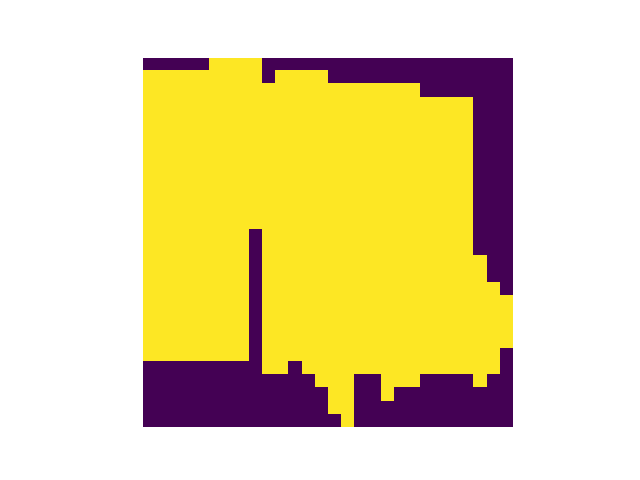}
}
\subfloat[Reconstructed]{
\includegraphics[width=0.24\linewidth]{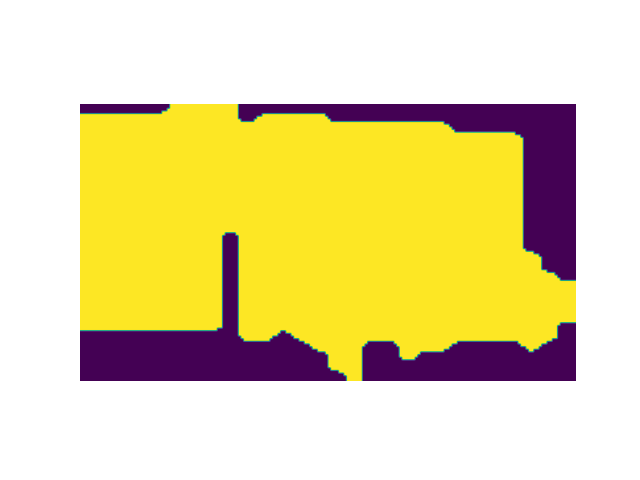}
}
\subfloat[Error]{
\includegraphics[width=0.24\linewidth]{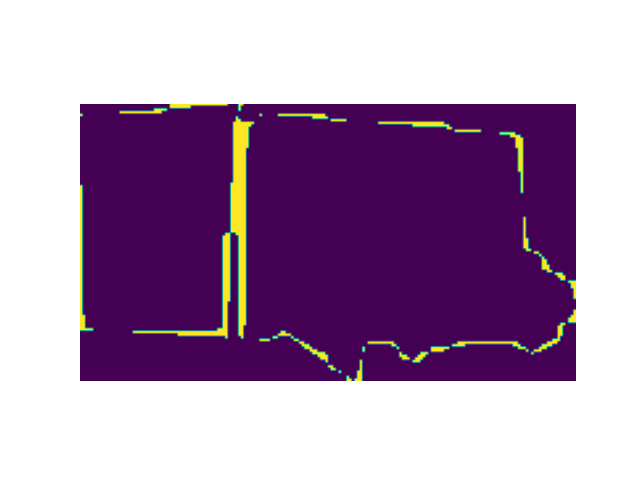}
}
\end{minipage}}}\\
\vspace{-7mm}
\scalebox{1.0}{
\subfloat[(b) DCT mask representation]{
\begin{minipage}[t]{0.9\linewidth}
\captionsetup[subfloat]{labelformat=empty}
\centering
\subfloat[Ground Truth]{
\includegraphics[width=0.24\linewidth]{figures/mask_source.png}
}
\subfloat[$128\times 128$]{
\includegraphics[width=0.24\linewidth]{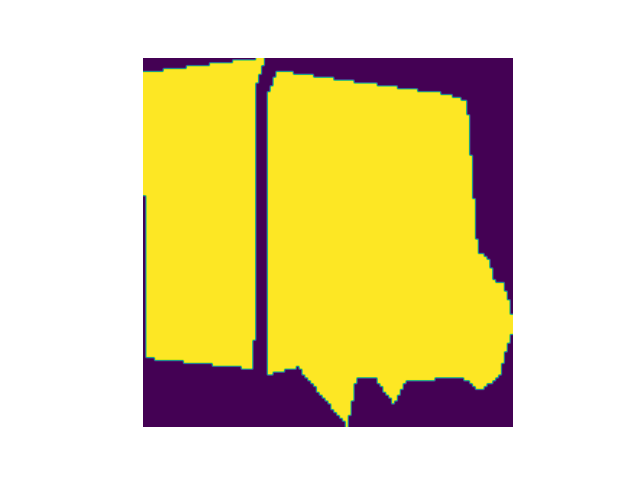}
}
\subfloat[Reconstructed]{
\includegraphics[width=0.24\linewidth]{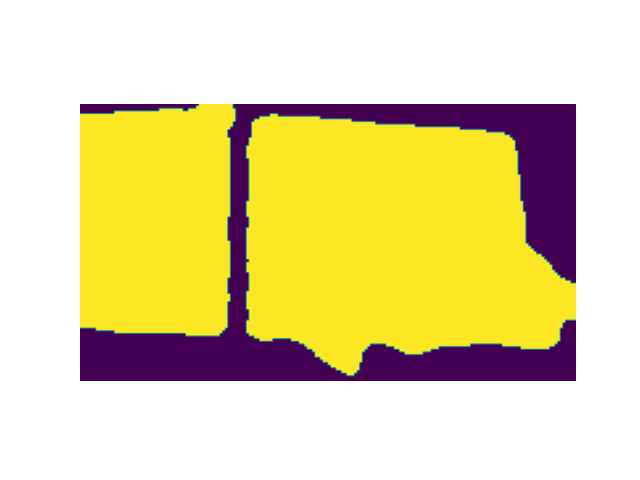}
}
\subfloat[Error]{
\includegraphics[width=0.24\linewidth]{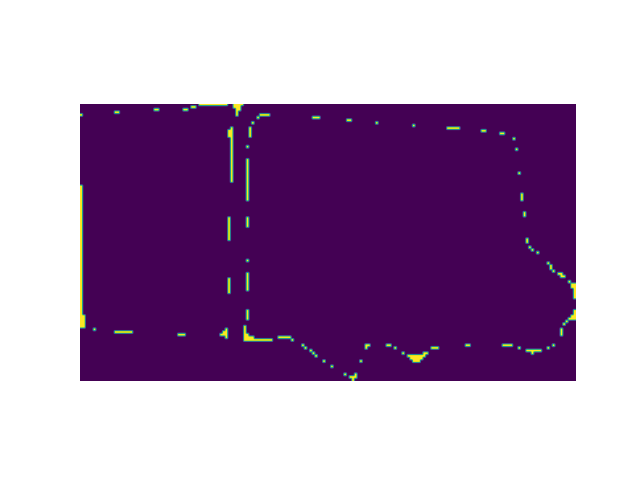}
}
\end{minipage}}}\\
\vspace{-2mm}
\caption{ \textbf{Binary grid mask representation vs. DCT mask representation.} The leftmost sub-graph is the ground truth, the left-middle is the mask representation, the middle right is the reconstructed mask, the rightmost is the error between the reconstructed mask and the ground truth.
}
\label{fig:mask_error}
\vspace{-2mm}
\end{figure}

{Instance segmentation tasks involve detecting objects and assigning category labels to pixels.  It is the cornerstone of many computer vision tasks, such as autonomous driving and robot manipulation.  Recently, the application of deep convolutional neural networks(CNNs) has greatly promoted the development of instance segmentation \cite{li2017fully,liu2018path,he2016deep,he2017mask,huang2019mask}. Pixel-based method is one of the mainstream methods which generates a bounding box by an object detector and performs pixel mask predicting within a low-resolution regular grid. But is the low-resolution grid an ideal choice for mask representation? 

As a representative instantiation in instance segmentation, Mask R-CNN \cite{he2017mask}} first downsamples the binary ground truth to a $28\times 28$ grid and then reconstructs it through upsampling. As shown in Figure \ref{fig:mask_error}(a), the binary grid mask representation with low resolution is not sufficient to capture detailed features and causes bias during upsampling. To describe this phenomenon more accurately, we use the metric of Intersection over Union(IoU) between the reconstructed mask and the ground truth to evaluate the quality of mask representation. We find that in the COCO \emph{val2017} dataset,
 the $28\times 28$ binary grid achieves only 93.8\% IoU. It means even the prediction of the mask branch is exactly accurate, the reconstructed mask has a 6.2\% error.

A higher-quality mask representation may reduce the reconstruction error. It turns out $128\times 128$ resolution could achieve 98\% IoU. But experiments show that the mask average precision(AP) of predicting a higher resolution binary grid is worse than the original $28\times 28$ resolution. (The specific experimental results will be discussed in the Experiments section.) The discriminative pixels of the mask distribute along the object boundaries, while the binary grid mask representation makes excessive predictions over all pixels on the whole large grid.
The training complexity will increase sharply as the resolution increases. The improvement in the quality of mask representation can not compensate for its shortcoming. 

A better mask representation with high resolution and low complexity is required. In this work, we explore to apply the discrete cosine transform to the mask representation. The discrete cosine transform (DCT) \cite{britanak2010discrete} is a widely used transformation technique in signal processing and data compression due to its strong ``energy compaction'' property \cite{rao2014discrete}. 
%For strongly correlated Markov processes, the DCT can approach the compaction efficiency of the Karhunen-Loeve transform which is optimal in the decorrelation sense. %!!
Regarding mask in instance segmentation as a binary image, most information is concentrated in a few low-frequency components. %and some high-frequency components are negligible. 
By transforming the high-resolution binary mask into the frequency domain and keeping its low-frequency components, we obtain a high-quality and low-complexity mask representation, termed DCT mask representation. Experiments indicate that it achieves 97\% IoU with only a 300-dimensional vector compressed from a $128\times 128$ resolution mask. The information of the discarded high-frequency components is negligible compared with the improvement brought by higher resolution. Consequentially the overall quality of mask representation is improved. Different from dictionary learning compression methods such as principal component analysis(PCA), sparse coding, and autoencoders, DCT does not require any pre-training or preprocessing, and the computational complexity of DCT transformation is $O(nlog(n))$ which could be omitted in the whole framework of instance segmentation. 

In this paper, we integrate the DCT mask representation into the Mask R-CNN \cite{he2017mask} framework with slight modifications to the model architecture. 
%different  pixel-based frameworks including Mask R-CNN \cite{he2017mask},  
Our method termed DCT-Mask consistently improves mask AP by about 1.3\% on the COCO dataset, 2.1\% on the LVIS\footnote{We use the COCO sub-categories of the LVIS dataset and evaluate it with same models trained on COCO.} dataset, and 2.0\% on the Cityscapes. Because DCT mask representation raises the upper bound of the mask quality, it tends to obtain more increase for more complex backbones and higher-quality annotations. On the COCO dataset, DCT-Mask achieves 1.3\% and 1.7\% increase with ResNet-50\cite{he2016deep} and ResNeXt-101\cite{xie2017aggregated}, respectively. On the LVIS$^*$ dataset , it achieves 2.1\% and 3.1\% increase respectively. Besides, we demonstrate the generality of our method on other pixel-based instance segmentation frameworks such as CascadeRCNN \cite{cai2018cascade}. %and MEInst \cite{zhang2020mask}.

DCT-Mask does not require extra pre-processing or pre-training. Compared to the standard Mask R-CNN with ResNet-50 backbone, 
the application of our method shows almost no harm to the running speed, which reaches 22 FPS on the V100 GPU. %and it only takes 8 hours to train 12 epochs on the COCO \emph{2017trainval} dataset.
Besides, we analyze the performance of mask prediction from the perspective of the mask quality and reveal that at the same level of complexity, improving the quality of mask representation can effectively improve mask AP. 

In summary, this work has the following contributions:
\begin{itemize}
\item We propose a high-quality and low-complexity mask representation for instance segmentation, which encodes the high-resolution binary mask into a compact vector with discrete cosine transform. 

\item With slight modifications, DCT-Mask could be integrated into most pixel-based frameworks, and achieve significant and consistent improvement on different datasets, backbones, and training schedules. Specifically, it obtains more improvements for more complex backbones and higher-quality annotations.

\item DCT-Mask does not require extra pre-processing or pre-training. It achieves high-resolution mask prediction at a speed similar to low-resolution.
\end{itemize}

\section{Related Works}
\textbf{Discrete cosine transform in computer vision.}
Discrete cosine transform is widely used in classic computer vision algorithms \cite{chadha2011face,ravi2016semantic,wei2002image} which encodes the spatial-domain RGB images into components in the frequency domain. With the development of deep learning, several studies investigate to integrate the DCT into computer vision frameworks based on deep learning. Ulicny \emph{et al.} \cite{ulicny2017using} used a CNN to classify DCT encoded images. Ehrlich \emph{et al.}  \cite{ehrlich2019deep} proposed a DCT-domain ResNet. Lo \emph{et al.} \cite{lo2019exploring}
 performs semantic segmentation on the DCT representation by feeding the rearranged DCT coefficients to CNNs. Xu \emph{et al.} \cite{xu2020learning} explores learning in the frequency domain for object detection and instance segmentation, which uses DCT coefficients as the input of CNN models instead of the RGB input. In these works, DCT is utilized to extract features of the model input. Here, we apply the DCT to represent the ground truth of mask in instance segmentation.

% or pixel based ? 
\textbf{Mask representation in instance segmentation.} Mask representation is the cornerstone of instance segmentation. Pixel-based instance segmentation methods \cite{dai2016instance,he2017mask,huang2019mask,liu2018path} represent the mask of objects on the pixel level within a region proposal. A representative instantiation is Mask R-CNN which predicts masks on a $28\times 28$ grid irrespective of object size. As discussed before, the low-resolution grid mask representation is low-quality, and the high-resolution grid suffers from high complexity. 

Several studies investigate to improve mask quality. % 这里应该还是用mask representation 与后边的对应，
Cheng \emph{et al.} \cite{cheng2020boundary} proposed boundary-preserving mask head
which aligns the predicted masks with object boundaries.
Mask Scoring R-CNN \cite{huang2019mask} proposed MaskIoU head to learn the quality of the predicted instance masks. It calibrates the misalignment between mask quality and mask score.
PointRend \cite{kirillov2020pointrend} regards image segmentation as a rendering problem, and obtains high-resolution segmentation predictions by iteratively refining the low resolution predicted mask at adaptively selected locations. It achieves $224\times 224$ resolution by 5 iterations starting from $7\times 7$ which yields significantly detailed results. But multiple iterations also increase the inference time. % These researches improve the performance of instance segmentation by a high quality mask, but also reduce the running speed.

Other studies investigate to reduce the complexity of mask representation. Jetley \emph{et al.} \cite{jetley2017straight} uses a denoising convolutional auto-encoder to learn a low-dimensional shape embedding space, and regresses directly to the encoded vector by a deep convolutional network. This shape prediction approach is hard to cope with a larger variety of shapes and object categories. MEInst\cite{zhang2020mask} uses PCA to encode the two-dimensional mask into a compact vector and incorporates it into a single-shot instance segmentation framework. But it only encodes $28\times 28$ resolution mask, and performs poorly on large objects due to the low quality of mask representation. PolarMask \cite{xie2020polarmask} represents the mask by its contour in the polar coordinates, and formulates the instance segmentation problem as predicting contour of instance through instance center classification and dense distance regression in a polar coordinate. %\cite{peng2020deep} proposed a learning-based snake algorithm which deforms an initial contour to the object boundary and introduces the circular convolution for feature learning on the contour. deepsnake方法的结果比baseline好，放到这里不合适
These studies achieve higher running speed with more compact mask representations. But their mask quality deteriorates, and the performance of instance segmentation is not ideal.

In this paper, we explore to improve mask quality as well as reducing the complexity to achieve a balance between performance and running speed.

\begin{table}[]
\centering
\scalebox{0.95}{
\begin{tabular}{c|c|c}
\Xhline{2\arrayrulewidth}
Resolution & $AP$  & IoU   \\ \hline
$28\times 28$      & 35.2 & 0.938 \\ 
$64\times 64$      & 34.4 & 0.968 \\ 
$128\times 128$    & 32.9 & 0.98  \\ \Xhline{2\arrayrulewidth}
\end{tabular}}
\vspace{-2mm}
\caption{Mask AP of Mask R-CNN with different resolution mask grid. Directly increasing the resolution decreases the mask AP.}
\vspace{-2mm}
\label{table: maskrcnn_high}
%很多文章都在表格中直接加上结论。
\end{table}

\section{Method}
As shown in Table \ref{table: maskrcnn_high}, when the resolution of binary mask representation increases from $28\times 28$ to $128\times 128$, the mask quality is improved, and the output size of mask branch increases from $784$ to 16384. Suffering from the massive increase of training complexity, the mask AP significantly decreases. Here, we propose DCT mask representation to reduce complexity.
% 它占用了很大的内存，且提高了很大的训练复杂度，但是maskAP 却下降了

\begin{figure}[]
\begin{center}
%\fbox{\rule{0pt}{2in} \rule{0.9\linewidth}{0pt}}
   \includegraphics[width=1.0\linewidth]{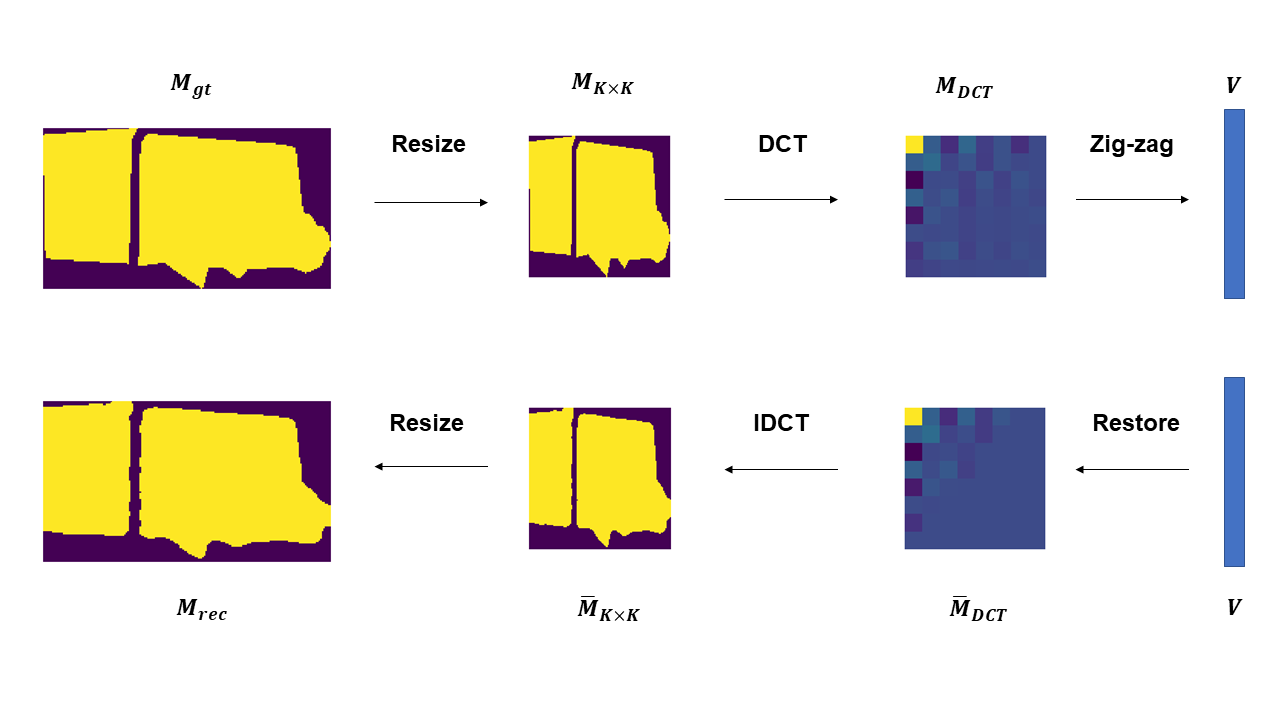}
\end{center}
\vspace{-7mm}
   \caption{The pipeline of DCT mask representation.}
\vspace{-2mm}
\label{fig:dct_pipeline}
\end{figure}

\subsection{DCT mask representation}
Our method is motivated by the JPEG standard \cite{wallace1992jpeg}, an  image file format. The pipeline of DCT mask representation is similar to JPEG which encodes the binary mask into a compact vector. As shown in Figure \ref{fig:dct_pipeline}, for binary ground truth mask $M_{gt}\in R^{H\times W}$, where H and W denotes the height and width, respectively. We resize it into $M_{K\times K}\in R^{K\times K}$ with bilinear interpolation where $K\times K$ is the mask size. Throughout this paper, $K=128$. It should be noted that $K=28$ in Mask R-CNN.

Two-dimensional DCT-II transforms $M_{K\times K}$ to the frequency domain $M_{DCT} \in R^{K\times K}$:
\begin{align}
M&_{DCT}(u, v)=\frac{2}{K} C(u) C(v) 
\notag
\\ &\sum_{x=0}^{K-1} \sum_{y=0}^{K-1} M_{K\times K}(x, y) \cos \frac{(2 x+1) u \pi}{2 K} \cos \frac{(2 y+1) v \pi}{2 K},
\label{eq: dct}
\end{align}
where $C(w) = 1/\sqrt{2}$ for $w=0$ and $C(w)=1$ otherwise.

Because of the strong ``energy compaction'' property of DCT, we select the first N-dimensional vector $V\in R^{N}$ from $M_{DCT}$ in a ``zig-zag''. Here the DCT mask vector $V$ is the mask representation we desire.

We restore $\bar{M}_{DCT} \in R^{K\times K}$ from $V$ by filling in the other parts with $0$. The next step is to take the two-dimensional inverse DCT (a 2D type-III DCT):
\begin{align}
\bar{M}_{K\times K}(x, y)=&\frac{2}{K} \sum_{u=0}^{K-1} \sum_{v=0}^{K-1} C(u) C(v) \bar{M}_{DCT}(u, v) 
\notag
\\ &\cos \frac{(2 x+1) u \pi}{2 K} \cos \frac{(2 y+1) v \pi}{2 K},
\end{align}
Finally, the bilinear interpolation is adopted to resize $\bar{M}_{K\times K}$ into $M_{rec}\in R^{H\times W}$.

From the above, we encode the ground truth of mask $M_{gt}$ as a compact vector $V$, then decode $V$ to reconstruct the mask $M_{rec}$. In this way, we use an N-dimensional vector $V$ as mask representation instead of a binary image which significantly reduces the redundancy. From Figure \ref{fig:dct_pipeline}, DCT mask representation captures the shape of the object accurately, and the discarded high-frequency component is only a few points from the boundary. 

As shown in Table \ref{table: dct_iou}, We evaluate the quality of mask representation by the metric of IoU between $M_{gt}$ and $M_{rec}$. DCT mask representation uses 100-dimensional and 700-dimensional vectors to achieve the same IoU as the $28\times 28$ and $128\times 128$ matrices in binary grid mask representation. This shows the efficiency of DCT mask representation. Because the size of most objects is smaller than $256\times 256$ on COCO, the results of $256\times 256$ are close to $128\times 128$. Therefore, $128\times 128$ resolution mask is sufficient for the task of instance segmentation. % Note that different interpolation methods may affect the value of IoU, but DCT mask representation still achieves higher mask quality than the binary mask grid.
% 其他文章也没提到插值的问题，前面也基本没提插值。

%Moreover, DCT and IDCT are essentially linear transformations. 
Moreover, fast cosine transform (FCT) algorithm \cite{haque1985two} computes DCT operations with only $O(nlog n)$ complexity, so the amount of calculation in this part is negligible. DCT can be perfectly integrated into the deep learning framework for training and inference.

In summary, without any pre-processing or pre-training, DCT mask representation effectively captures the details of a mask with low complexity. It casts the task of mask prediction into regressing the DCT vector $V$.

\begin{table}
\centering
\scalebox{0.95}{
\begin{tabular}{c|c|c}
\Xhline{2\arrayrulewidth}
 Resolution & Dim & IoU \\ \hline
 $28 \times 28$ & None &  0.938 \\
 $64 \times 64$ & None &  0.968 \\
 $128 \times 128$ & None &  0.980 \\
\hline $128 \times 128$ & 100 &  0.940 \\
 $128 \times 128$ & 300 &  0.970 \\
 $128 \times 128$ & 500 &  0.976 \\
 $128 \times 128$ & 700 &  0.979 \\
 $256 \times 256$ & 100 &  0.940 \\ 
 $256 \times 256$ & 300 &  0.970 \\ 
 $256 \times 256$ & 500 &  0.977 \\ 
 $256 \times 256$ & 700 &  0.980 \\  
\Xhline{2\arrayrulewidth}
\end{tabular}}
\vspace{-2mm}
\caption{The quality of DCT mask representation with different dimensions and resolution. ``None'' stands for the binary grid mask representation.}
\vspace{-2mm}
\label{table: dct_iou}
\end{table}
\begin{figure}[t]
\centering
\subfloat[Mask R-CNN]{
\begin{minipage}[t]{0.8\linewidth}
\centering
\includegraphics[width=0.8\linewidth]{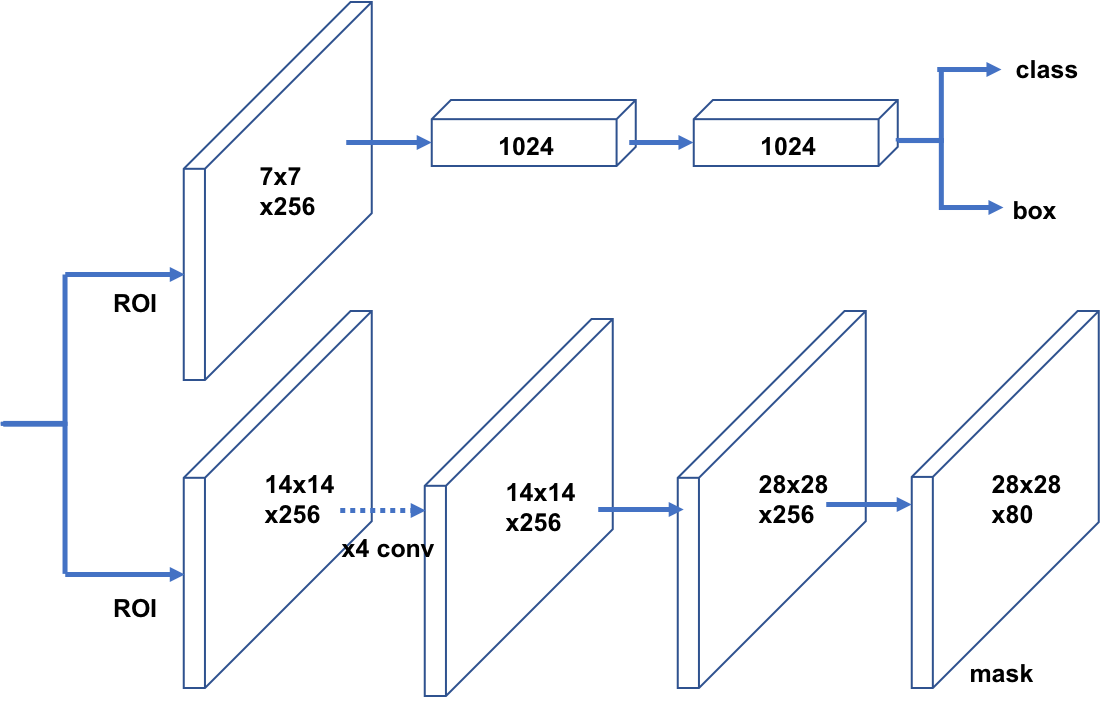}
\end{minipage}
}\\
\subfloat[DCT-Mask R-CNN]{
\begin{minipage}[t]{0.8\linewidth}
\centering
\includegraphics[width=0.8\linewidth]{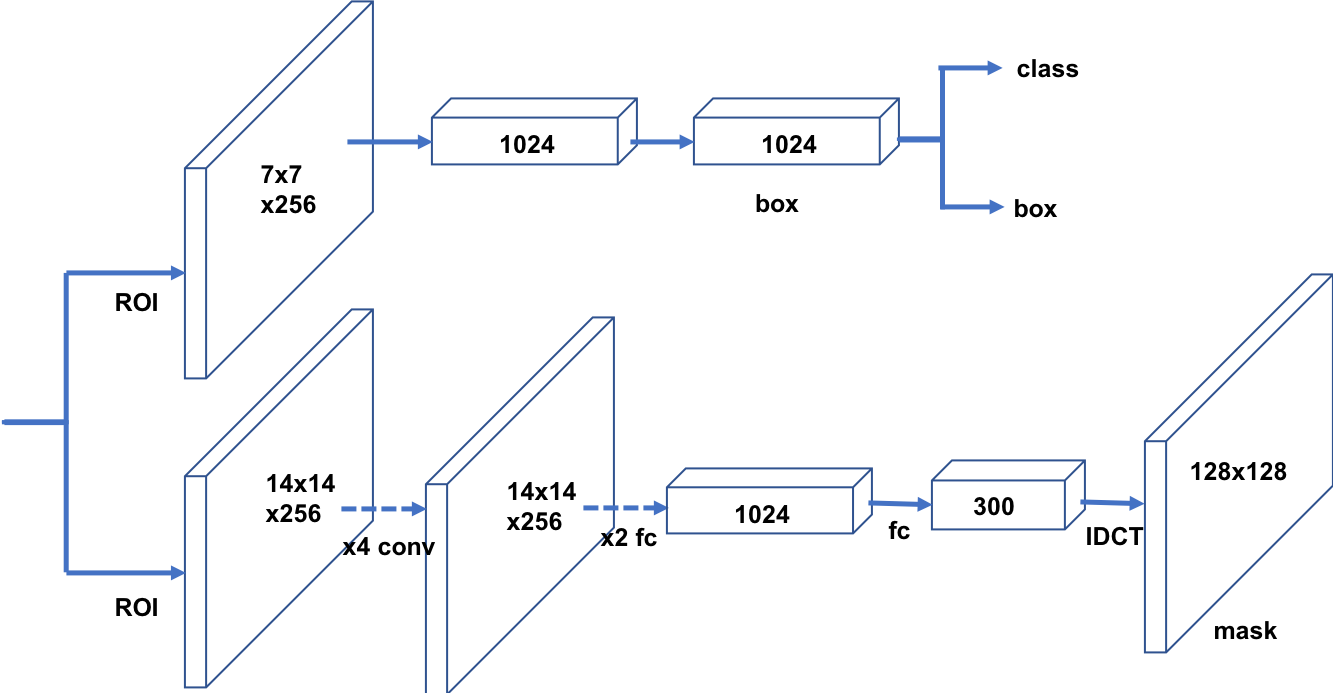}
\end{minipage}
}
\vspace{-2mm}
\caption{The implementation of DCT-Mask on the basis of Mask R-CNN.}
\vspace{-2mm}
\label{fig:maskrcnn}
\end{figure}

\subsection{DCT-Mask in Mask R-CNN}
We could integrate the DCT-mask into most pixel-based instance segmentation frameworks with slight modifications to the model architecture. In this section, we take Mask R-CNN as an example.

We begin by briefly reviewing the Mask R-CNN \cite{he2017mask}. Mask R-CNN is a two-stage instance segmentation method. The first stage generates proposals about the regions by the Region Proposal Network(RPN). The second stage consists of detection branch and mask branch. The detection branch predicts the class of the object and refines the bounding box based on the first stage proposal by R-CNN Head. The mask branch performs pixel classification to generate the mask of the object by mask head. 

By applying the DCT mask representation, the prediction of the mask branch is a compact vector instead of a binary grid. %Because the detection branch in Mask R-CNN regresses a vector for bound box and classification respectively, we imitate the architecture of the detection branch. 
As shown in Figure \ref{fig:maskrcnn}, we use 4 convolution layers to extract the feature of mask, and 3 fully connected layers to regress the DCT mask vector. The setting of convolution layers is the same as Mask R-CNN where the kernel size and filter number are 3 and 256 respectively. The outputs of the first two fully-connected layers have the size of 1024, and the output size of the last layer is the dimension of the DCT mask vector. Furthermore, the prediction of mask head is class-agnostic which reduces the training complexity by keeping a small output size.

With DCT mask representation, the ground truth of mask branch is a vector encoded by DCT. It leads to a regression problem. Then we define the loss function of mask branch as following: 
\begin{equation}
\mathcal{L}_{mask} = \mathtt{1}^{obj}\sum_i^N D(\hat{V}_i,V_i) \space ,
\end{equation}
where $V_i,\hat{V_i}$ denotes $i$-th element in ground-truth and prediction vectors, respectively. $\mathtt{1}^{obj}$ is the indicator function for positive samples. $D$ is the distance metric which is $l_1$ loss in this article.

We define the loss function of the whole model,
\begin{equation}
\mathcal{L}= \mathcal{L}_{{det}}+\lambda_{{mask}} \cdot \mathcal{L}_{{mask}} \space ,
\end{equation}
where $\mathcal{L}_{det}$ is the loss for the detection branch. $\lambda_{mask}$ is the corresponding weight. 

During inference, we follow the standard Mask R-CNN inference procedure. After NMS, top-k scoring boxes are selected and fed into the Mask branch with RoIAlign. Mask branch predicts a DCT mask vector for each bounding box. The mask within the box is generated by decoding the DCT mask vector. 

In summary, keeping the other parts completely unchanged, we only modify the mask branch by replacing the last 2 layers with 3 fully connected layers. Similarly, our method can be easily applied to other pixel-based instance segmentation frameworks 

\section{Experiments}
\subsection{Datasets}
We evaluate our method on COCO \cite{lin2014microsoft} and Cityscapes datasets \cite{cordts2016cityscapes}. We use COCO evaluation metrics AP (averaged over IoU thresholds) to report the results including AP@0.5, AP@0.75, $AP_S$, $AP_M$, and $AP_L$ (AP at different scales).

COCO has 80 categories with instance-level annotation. Following COCO 2017 settings, we train on \emph{train2017} (about 118k images) and validate
on \emph{val2017} (about 5k images).  The COCO ground-truth is often too coarse to reflect improvements in mask AP while LVIS \cite{gupta2019lvis} re-annotates the COCO dataset with higher quality masks. %Following the approach of \cite{kirillov2020pointrend}
We use the same models trained on the COCO dataset, and re-evaluate it on the COCO category subset of LVIS, denoted as LVIS$^*$.
% can more realistically evaluating mask improvements 
%LVIS is a high-quality instance segmentation dataset with higher-quality annotations which has 2.2 million instance segmentation masks. Note that the same model trained on COCO is used to re-evaluate on LVIS. 

Cityscapes is a widely used ego-centric street-scene dataset with 2975 training, 500 validation, and 1525 testing images with high-quality annotations. We evaluate the performance in terms of the average precision (AP) metric averaged over 8 semantic classes of the dataset.
 %Each image is with size
%1024  2048.
%Compared to COCO, it has high-quality annotations and higher resolution.  

\subsection{Implementation Details}
In this paper, we choose $128\times 128$ mask size and 300-dimensional DCT mask vector as default mask representation. Our method is implemented on both Detectron2 \cite{wu2019detectron2} and MMDetection \cite{mmdetection}, the experimental results of these two toolboxes 
are consistent. 
We use the standard $1\times $ training schedule and multi-scale 
training from Detectron2 by default. On COCO, the experiments adopt 
90K iterations, batch size 16 on 8 GPUs, and base learning rate 0.02. 
The learning rate is reduced by a factor of 10 at iteration 60K and 
80K. Multi-scale training is used with shorter side randomly sampled 
from $[640,800]$. The short side is resized to 800 in inference. On Cityscapes, we use 24K iterations and 0.01 base 
learning rate, and reduce the learning rate at 18K. Shorter side is 
randomly sampled from $[800,1024]$ in training, and resized to 1024 in inference.
%The experiments with ResNet-101 and ResNeXt-101 backbones are trained with the $3\times$ schedule(270K iterations). 
%blend mask
%Code will be available soon. 

\textbf{Loss function and corresponding weight.}
\begin{figure}
\centering
\subfloat[Mean]{
\includegraphics[width=0.2\textwidth]{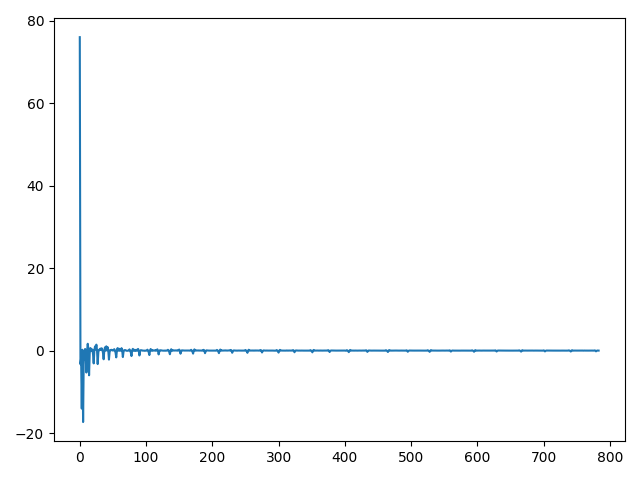}
}
\subfloat[Variance]{
\includegraphics[width=0.2\textwidth]{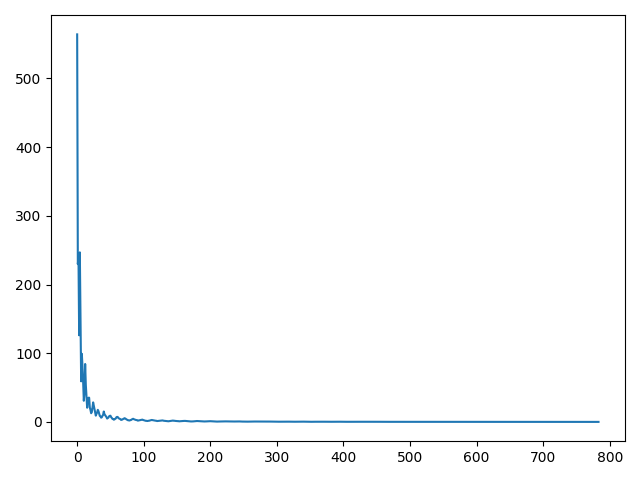}
}
\vspace{-2mm}
\caption{Mean value and Variance of DCT mask vectors on COCO 2017 val dataset.}
\vspace{-2mm}
\label{fig:meanvar}
\end{figure}
As shown in Figure \ref{fig:meanvar},
because of the energy concentration characteristics of DCT, most information concentrates on the previous dimensions. The mean and variance of previous dimensions are much larger than the latter. Hence $l_2$ loss would be unstable in the training process. In Table \ref{table:loss_weight},  we compare $l_1$ loss and smooth $l_1$ loss with different corresponding weights $\lambda_{mask}$. It turns out $l_1$ and smooth $l_1$ have similar performance, and DCT-Mask is robust to the corresponding weight. We choose the best combination of $l_1$ loss and $\lambda_{mask}=0.007$ in this paper. 
%Moreover, from Equation \ref{eq:dct} we can derive that the first item of the DCT coefficients is related to the mask size $K$. 
%Double the mask size and the first term will double, the corresponding weight should be halved. For example, we use $\lambda_{mask}=0.0035$ for $256\times 256$ resolution. 
% 深度学习中一般默认l2 loss = |x-y|^2, l1 loss = |x-y|

\begin{table}[]
\centering
\scalebox{0.95}{
\begin{tabular}{c|c|c|c}
\Xhline{2\arrayrulewidth}
Loss                 & Weight & COCO AP & LVIS$^*$ AP \\ \hline
\multirow{4}{*}{$l_1$}  & 0.005  & 36.2    & 39.3    \\ %\cline{2-4} 
                     & 0.006  & 36.4    & 39.4    \\ %\cline{2-4} 
                     & 0.007  & 36.5    & 39.6    \\
                     & 0.008  & 36.2    & 39.2    \\
                      \hline
\multirow{4}{*}{smooth $l_1$} 
					& 0.004  & 36.4    & 39.1    \\
					& 0.005  & 36.4    & 39.2    \\ %\cline{2-4} 
                     & 0.006  & 36.4    & 39.5    \\ %\cline{2-4} 
                     & 0.007  & 36.4    & 39.3    \\ \Xhline{2\arrayrulewidth}
\end{tabular}}
\vspace{-1mm}
\caption{The results of $l_1$ loss and smooth $l_1$ loss with different corresponding weights. The resolution is $128\times 128$, and the number of dimensions is 300.}
\vspace{-1mm}
\label{table:loss_weight}
\end{table}

\begin{table*}[]
\scalebox{0.85}{
\begin{tabular}{c|cccccc|cccccc|c}
\Xhline{2\arrayrulewidth}
\multirow{2}{*}{Method} & \multicolumn{6}{c|}{COCO}               & \multicolumn{6}{c|}{LVIS$^*$}               & Cityscapes \\ 
                        & AP   & AP@50 & AP@75 & AP$_S$  & AP$_M$  & AP$_L$  & AP   & AP@50 & AP@75 & AP$_S$  & AP$_M$  & AP$_L$  & AP         \\ \hline
Mask R-CNN                & 35.2 & 56.3 & 37.5 & 17.2 & 37.7 & 50.3 & 37.6 & 59.2 & 39.2 & 21.2 & 43.7 & 55.1 & 33.0       \\ 
DCT-Mask R-CNN            & 36.5 & 56.3 & 39.6 & 17.7 & 38.6 & 51.9 & 39.7 & 59.7 & 42.1 & 23.5 & 46.5 & 58.5 & 35.6       \\ \Xhline{2\arrayrulewidth}
\end{tabular}}
\vspace{-1mm}
\caption{Mask AP on the validation set of COCO, LVIS$^*$ and Cityscapes. ResNet-50 backbone and 1x training schedule are used. The results show that DCT-Mask yields higher AP gains with higher quality annotations.}
\vspace{-2mm}
\label{table: datasets}
\end{table*}

\begin{table*}[]
\scalebox{0.95}{
\begin{tabular}{l|l|l|lcc|lcc}
\Xhline{2\arrayrulewidth}
\multicolumn{1}{c|}{\multirow{2}{*}{Method}}                  & \multirow{2}{*}{Backbone}    & \multirow{2}{*}{DCT-Mask} & \multicolumn{3}{c|}{COCO} & \multicolumn{3}{c}{LVIS$^*$} \\
                                                  &                              &                           & AP      & AP@50   & AP@75   & AP      & AP@50   & AP@75   \\ \hline
\multicolumn{1}{c|}{\multirow{6}{*}{Mask R-CNN}} & \multirow{2}{*}{ResNet-50}   &                           & 35.2    & 56.3   & 37.5   & 37.6    & 59.2   & 39.2   \\
\multicolumn{1}{c|}{}                            &                              & \multicolumn{1}{c|}{\checkmark}    & 36.5(+1.3)    & 56.3   & 39.6   & 39.7(+2.1)    & 59.7   & 42.1   \\ \cline{2-9} 
\multicolumn{1}{c|}{}                            & \multirow{2}{*}{ResNet-101}  &                           & 38.6    & 60.4   & 41.3   & 41.4    & 63.0   & 44.0   \\
\multicolumn{1}{c|}{}                            &                              & \multicolumn{1}{c|}{\checkmark}    & 39.9(+1.3)    & 60.5   & 43.4   & 43.7(+2.3)    & 63.6   & 46.8   \\ \cline{2-9} 
\multicolumn{1}{c|}{}                            & \multirow{2}{*}{ResNeXt-101} &                           & 39.5    & 61.7   & 42.6   & 42.1    & 63.6   & 45.0   \\
\multicolumn{1}{c|}{}                            &                              & \multicolumn{1}{c|}{\checkmark}    & 41.2(+1.7)    & 62.4   & 44.9   & 45.2(+2.9)    & 65.7   & 48.5   \\ \hline
\multirow{6}{*}{Casecade Mask R-CNN}              & \multirow{2}{*}{ResNet-50}   &                           & 36.4    & 56.9   & 39.2   &  38.9       &  59.9      &   41.2     \\
                                                  &                              & \multicolumn{1}{c|}{\checkmark}    & 37.5(+1.1)    & 57.0   & 40.8   &  40.9(+2.0)       &  60.1      &   43.5     \\ \cline{2-9} 
                                                  & \multirow{2}{*}{ResNet-101}  &                           &  39.6       &   60.9     &  42.9      &   42.2      &   63.5     &   45.3     \\
                                                  &                              & \multicolumn{1}{c|}{\checkmark}    &  40.8(+1.2)       &  61.5      &  44.4      &  44.3(+2.1)      &   63.8     &   47.8     \\ \cline{2-9} 
                                                  & \multirow{2}{*}{ResNeXt-101} &                           &  40.2       &  62.3      &   43.5     &   43.2      &  64.4      &  46.3      \\
                                                  &                              & \multicolumn{1}{c|}{\checkmark}    &   42.0(+1.8)      &    63.0    &   45.8     &   46.0(+2.8)      &  66.3      &    49.5    \\ \Xhline{2\arrayrulewidth}
\end{tabular}}
\vspace{-2mm}
\caption{Mask AP on COCO and LVIS$^*$ validation dataset with different backbones. The results without \checkmark are those of standard Mask R-CNN and Cascade Mask R-CNN, while with \checkmark are those of DCT-Mask. 1x training schedule is used for ResNet-50, 3x training schedule is used for ResNet-101 and ResNeXt-101. The results shows that our method yields higher AP gains with more complex backbones.}
\vspace{-2mm}
\label{table: backbones}
\end{table*}

\subsection{Main Results}
%Visual results are shown in Figure 5.

%We evaluate and express the effectiveness of our method on different backbone networks including ResNet-50 \cite{he2016deep}, ResNet-101 and ResNeXt-101 \cite{xie2017aggregated}, different frameworks including Mask R-CNN\cite{he2017mask},Cascade Mask R-CNN\cite{cai2018cascade} and MEInst\cite{zhang2020mask}, % and HTC \cite{chen2019hybrid} and different datasets including COCO and Cityscapes. 

We compare DCT-Mask R-CNN to the standard Mask R-CNN with ResNet-50 backbone in Table \ref{table: datasets}. On the COCO, LVIS$^*$, and Cityscapes datasets, mask AP of our method is increased by 1.3\%, 2.1\%, 2.6\% respectively. Especially for AP@75, DCT-Mask outperforms baseline by 2.1\% and 2.9\% on COCO and LVIS$^*$, respectively.
Recalling that Cityscapes and LVIS$^*$ have higher-quality annotations, which allows them to evaluate mask improvements in more details. DCT-Mask yields more increments in AP$_M$ and AP$_L$ than AP$_S$, because larger objects need higher-resolution mask representation to capture the details.
%Because of the high resolution mask representation, DCT-Mask yields more increasement in APm and APl than APs. %which stands for small objects with area less than $32^2$. 

%Besides, it achieves higher mask AP gains when evaluating the COCO categories with the LVIS annotations(LVIS$^*$) and for Cityscapes, which is attributed to the higher-quality annotations. 

DCT mask representation allows our method to yield higher AP with lower FLOPs. As shown in Table \ref{table: FPS}, compared to Mask R-CNN with $128\times 128$ resolution, our method brings a 3.6\% AP gain with more than 20 times less computation (FLOPs). Moreover, DCT-Mask predicts $128\times 128$ resolution masks at 22 FPS which is almost the same as Mask R-CNN with $28\times 28$ resolution. 
%we obtain $128\times 128$ resolution mask and higher mask AP,
 %and the same FLOPs and FPS as $28\times 28$ resolution in the Mask R-CNN.

Theoretically, DCT-Mask can be integrated into most pixel-based instance segmentation frameworks. At Table \ref{table: backbones}, we show that DCT-Mask achieves consistent performance on Cascade Mask R-CNN\cite{cai2018cascade}.  

It is worth to mention that our method has a greater improvement for large backbones.  
We show this property in Table \ref{table: backbones}. Compared to the standard Mask R-CNN, DCT-Mask respectively increases 1.3\% and 1.7\% mask AP with ResNet-50 and ResNeXt-101 on COCO \emph{2017 val} dataset. On LVIS$^*$ dataset, it increases 2.1\% and  3.1\% mask AP respectively. 
The same property is also presented on Cascade Mask R-CNN\cite{cai2018cascade}. DCT-Mask respectively increases 1.1\% and 1.8\% mask AP with ResNet-50 and ResNeXt-101 on COCO, while it increases 2.0\% and 2.8\% mask AP on LVIS$^*$. Because DCT mask representation captures more detailed features and improves the quality of the mask, a larger backbone can make full use of this advantage, thereby further improving the results. 

\begin{table}[]
\scalebox{0.95}{
\begin{tabular}{c|c|c|c|c}
\Xhline{2\arrayrulewidth}
Method         & Resolution & AP & FLOPs & FPS \\ \hline
Mask R-CNN     & $28\times 28$      & 35.2   & 0.5B  & 23  \\ 
Mask R-CNN     & $64\times 64$     & 34.4   & 2.7B  & 19    \\ 
Mask R-CNN     & $128\times 128$    & 32.9   & 11.0B   & 13    \\ 
DCT-Mask R-CNN & $128\times 128$    & 36.5   & 0.5B  & 22  \\ \Xhline{2\arrayrulewidth}
\end{tabular}}
\vspace{-1mm}
\caption{Mask head FLOPs and FPS for a $128\times 128$ resolution mask. The FPS index is measured on V100. DCT-Mask increases 3.6\% AP with 20 times less FLOPs. }
\vspace{-1mm}
\label{table: FPS}
\end{table}

%Bnenefiting from high resolution mask prediction and higher upper bound of mask quality, DCT-Mask tends to increase more mask AP for larger backbones. 

%(Other methods may be added).
%This demonstrates the broad applicability of our method. 如果其他实验的结果好的话。

\begin{figure*}[t]
\centering
\captionsetup[subfloat]{labelformat=empty}
\subfloat{
\begin{minipage}[t]{1.0\textwidth}
\captionsetup[subfloat]{labelformat=empty}
\centering
\subfloat{
\includegraphics[width=0.33\textwidth]{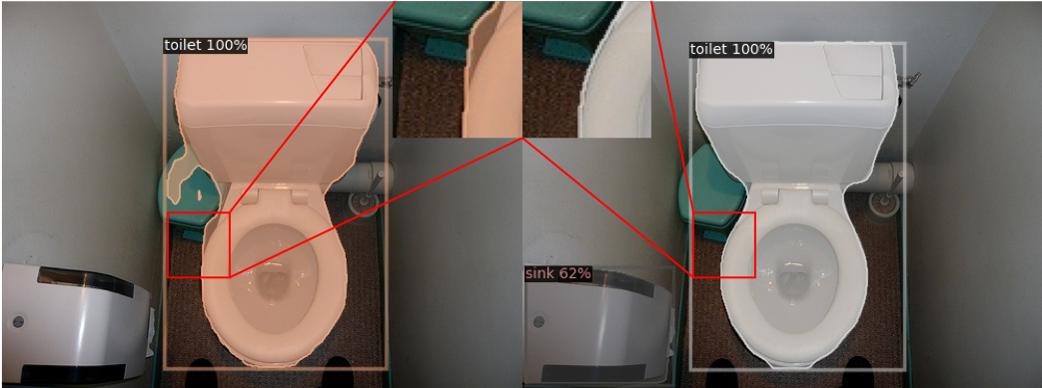}
}
\subfloat{
\includegraphics[width=0.33\textwidth]{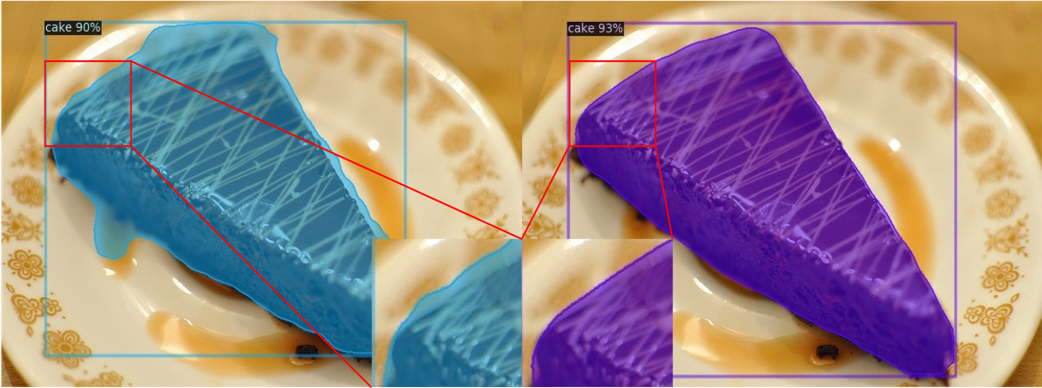}
}
\subfloat{
\includegraphics[width=0.33\textwidth]{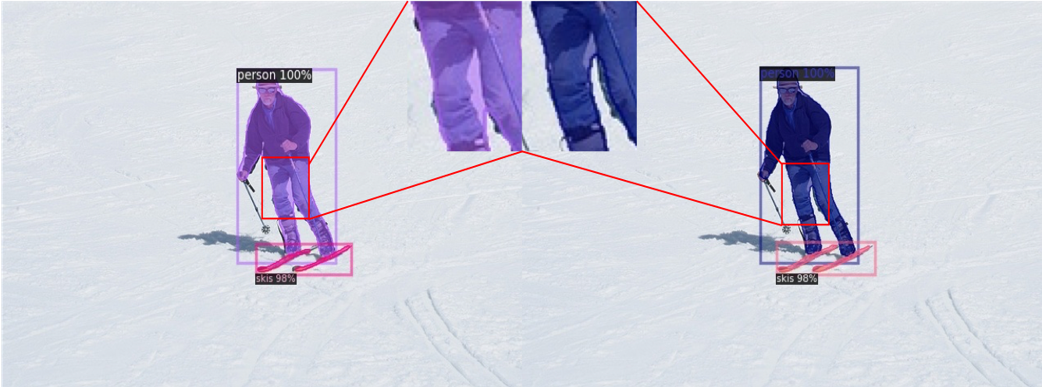}
}
\end{minipage}}\\
\vspace{-7mm}
\subfloat{
\begin{minipage}[t]{1.0\textwidth}
\captionsetup[subfloat]{labelformat=empty}
\centering
\subfloat{
\includegraphics[width=0.33\textwidth]{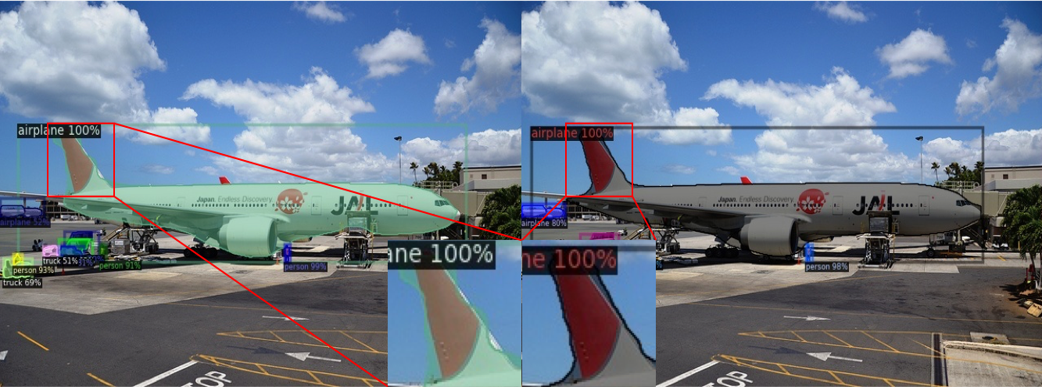}
}
\subfloat{
\includegraphics[width=0.33\textwidth]{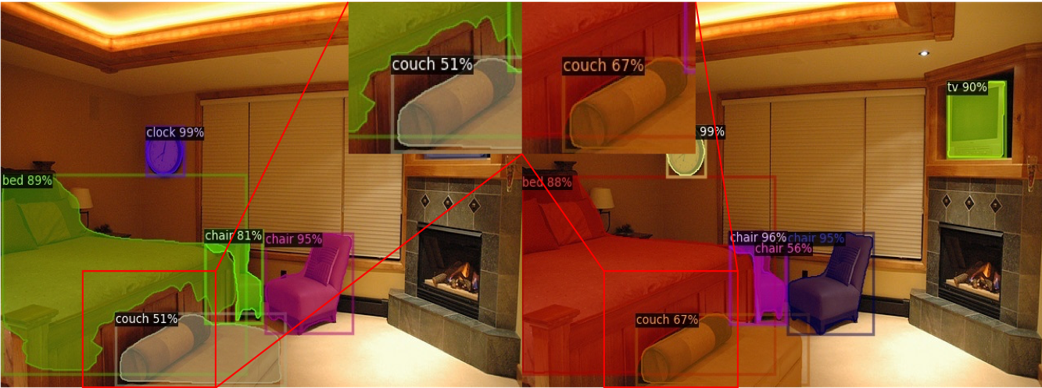}
}
\subfloat{
\includegraphics[width=0.33\textwidth]{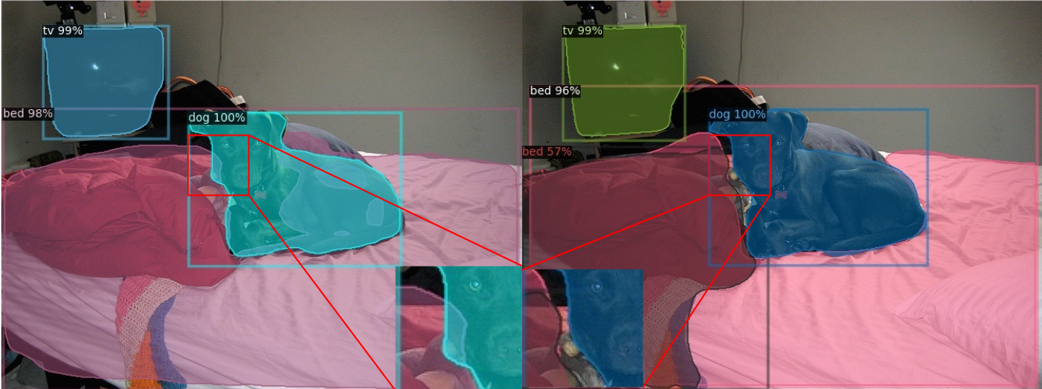}
}
\end{minipage}}\\
\vspace{-2mm}
\caption{Example result pairs from Mask R-CNN vs. with DCT-Mask R-CNN(right image), using ResNet-50 with FPN.}
\vspace{-2mm}
\label{fig: merge}
\end{figure*}

DCT-Mask R-CNN outputs are visualized in Figures \ref{fig: merge}. Compared to the standard Mask R-CNN, DCT-Mask obtains finer results around object boundaries. Moreover, we compare our method with other instance segmentation methods on COCO \emph{test-dev2017} and Cityscapes \emph{test} at Table \ref{table: sota} and Table \ref{table: cityscapes}. Without bells and whistles, DCT-Mask achieves state-of-the-art performances.

\subsection{Ablation Study}

\begin{table}[]
\centering
\scalebox{0.95}{
\begin{tabular}{c|c|c|c}
\Xhline{2\arrayrulewidth}
Resolution & Dim & AP  & IoU \\ \hline
$32\times 32$       & 300 &  35.4       &  0.950       \\ 
$64\times 64$       & 300 & 36.4  &  0.968       \\ 
$128\times 128$     & 300 & 36.5   &  0.970       \\ 
$256\times 256$     & 300 & 36.5        &  0.970       \\ \Xhline{2\arrayrulewidth}
\end{tabular}}
\vspace{-1mm}
\caption{DCT-Mask with different resolutions. It shows that higher resolution has higher IoU and mask AP.}
\vspace{-1mm}
\label{table: diff_resolution}
\end{table}

%As discussed before, we use the metric of IoU between $M_{gt}$ and $M_{rec}$ on COCO \emph{2017val} dataset to evaluate the quality of mask representation. 
In this part, we explore the reason why our method increases the mask AP from the perspective of mask quality and model architecture modifications. In these experiments, we use Mask R-CNN framework and ResNet-50 backbone, and evaluate on COCO \emph{2017val} dataset.

\textbf{Resolution of mask representation.} DCT mask representation can obtain high-quality mask with low complexity. As shown in Table \ref{table: diff_resolution}, when the dimension of the DCT vector is the same(Dim=300), as the resolution increases from $32\times 32$ to $128\times 128$, IoU increases from $0.95$ to $0.97$. DCT mask representation captures more information, and mask AP is increased from 35.4 to 36.5. When the resolution increases to $256\times 256$, the mask quality is almost unchanged, mask AP is the same as $128\times 128$.

\textbf{Dimension of DCT mask vector.} Because of the energy concentration characteristics of DCT, the ``importance'' of the latter dimension is less than the former. This is reflected in the experiments. As illustrated in Table \ref{table: diff_dim}, with the same resolution $128\times 128$, as the dimension increases from 100 to 900, the growth of mask quality is gradually slowed. When Dim $>$300, due to the increased training complexity, although the mask quality has improved, mask AP remains basically unchanged. %Considering performance and training complexity, we use $128\times 128$ resolution with a 300-dimensional DCT mask vector as default.

%\begin{table}[!htbp]
%\begin{tabular}{|l|l|l|l|}
%\hline
%resolution & Dim & $AP_m$  & IoU \\ \hline
%$28\times 28$ & None & 35.2 & 0.938 \\ \hline
%$28\times 28$ & 784 & 35.4 & 0.938 \\ \hline
%$32\times 32$       & 300 &  35.4       &  0.950       \\ \hline
%$64\times 64$ & None & 34.4 & 0.968 \\ \hline
%$64\times 64$       & 300 & 36.4  &  0.968       \\ \hline
%$128\times 128$ & None & 34.9 & 0.980 \\ \hline
%$128\times 128$     & 100 & 35.3        &  0.940       \\ \hline
%$128\times 128$     & 300 & 36.5   &  0.970       \\ \hline
%$128\times 128$     & 500 & 36.5        &  0.976       \\ \hline
%$128\times 128$     & 700 & 36.5        &  0.979       \\ \hline
%$256\times 256$     & 300 & 36.5        &  0.970       \\ \hline
%\end{tabular}
%\caption{Resolution and Dimension of DCT mask representation. "None" denotes the binary grid mask representation. }
%\label{resolution}
%\end{table}

\begin{table}[]
\centering
\scalebox{0.95}{
\begin{tabular}{c|c|c|c}
\Xhline{2\arrayrulewidth}
Resolution & Dim & $AP$  & IoU \\ \hline
$128\times 128$     & 100 & 35.3        &  0.940       \\ 
$128\times 128$     & 300 & 36.5   &  0.970       \\ 
$128\times 128$     & 500 & 36.5        &  0.976       \\ 
$128\times 128$     & 700 & 36.5        &  0.979       \\ 
$128\times 128$     & 900 & 36.4        &  0.980       \\ \Xhline{2\arrayrulewidth}
\end{tabular}}
\vspace{-2mm}
\caption{DCT mask vectors with different dimensions. As dimension increases, the growth of IoU is gradually slowed, and mask AP remains basically unchanged when $Dim>300$.}
\vspace{-2mm}
\label{table: diff_dim}
\end{table}

\textbf{Different representations with the same quality.}
For a 784-dimensional DCT mask vector of $28\times 28$ resolution, DCT could be regarded as a reversible transformation. It transforms the pixel mask representation into the frequency domain, and the quality of mask representation is the same. Besides, for a 100-dimensional DCT mask vector of $128\times 128$ resolution, the mask quality is $0.94$ which is close to binary mask representation with $28\times 28$ resolution in the standard Mask R-CNN. As shown in Table \ref{table: same_quality}, the performance of these experiments is very close to the Mask R-CNN baseline. This implies that the type of mask representation is not the reason for the increase in mask AP.

In summary of the previous three parts, DCT mask representation of different resolutions with the same mask quality has the same performance. Different types of mask representations with the same mask quality have the same performance. Besides, when mask quality increases as the dimension increases, the mask AP also increases. When mask quality remains basically unchanged as the dimension increases, the mask AP remains unchanged. We can conclude that the increase of mask AP is due to the increase of mask quality, not the type of mask representation. The reason why DCT-Mask works is that it achieves a balance between mask quality and complexity. 
% 这里进一步证明了结论
% 这里考虑更详细地进行描述。
%The perfor mance of our method is advanced by ....

\begin{table}[]
\centering
\scalebox{0.95}{
\begin{tabular}{c|c|c|c}
\Xhline{2\arrayrulewidth}
Resolution & Dim & AP  & IoU \\ \hline
$28\times 28$ & None & 35.2 & 0.938 \\ 
$28\times 28$ & 784 & 35.4 & 0.938 \\ 
$128\times 128$     & 100 & 35.3        &  0.940       \\ \Xhline{2\arrayrulewidth}
\end{tabular}}
\vspace{-2mm}
\caption{Different mask representations with similar mask quality. ``None'' denotes the binary grid mask representation in standard Mask R-CNN. Mask AP is close with similar mask quality.}
\vspace{-2mm}
\label{table: same_quality}
\end{table}

\textbf{The effect of the architecture modifications.}
\begin{table}[]
\centering
\scalebox{0.95}{
\begin{tabular}{c|cc|c}
\Xhline{2\arrayrulewidth}
\multicolumn{1}{l|}{Method} & 3 FC & L1 & AP   \\ \hline
\multirow{3}{*}{Mask R-CNN}  &      &    & 34.7 \\ 
                             & \checkmark    &    & 34.6 \\  
                             & \checkmark    & \checkmark  & 18.5 \\ \Xhline{2\arrayrulewidth}
\end{tabular}}
\vspace{-2mm}
\caption{The effect of architecture modifications. The added 3 FC layers shows negligible impact on the results. $l_1$ loss is not suitable for Mask R-CNN.}
\vspace{-2mm}
\label{table:fc_layers}
\end{table}
We integrate DCT mask representation into the standard Mask R-CNN with slight modifications to the model architecture. To make sure the improvement of our method is not due to these modifications, we add 3 fully connected(FC) layers and apply $l_1$ loss to the standard Mask R-CNN mask branch, respectively. Because the output size of the FC layers is 784, we use the class agnostic mask. As shown in Table \ref{table:fc_layers}, the impact of the FC layers could be ignored, and $l_1$ loss is not suitable for binary grid mask.

\textbf{The design of the mask branch.}
We investigate the impact of the detailed setting of mask branch. As we present before, we use 4 convolution layers and 3 FC layers in mask branch and achieves 36.5 AP on COCO with ResNet-50. We find that adding one more convolution layer or FC layer will achieve 36.4 AP and 36.3 AP respectively. Moreover, when we increase the hidden size of FC layers from 1024 to 2048, it achieves 36.5 AP. %It implies that DCT-Mask has stable performance on different settings of mask branch. 
%As shown in Table \ref{table:mask branch}, we evaluate the performance including adding one more convolution layer, adding one more fully-connected layer and increasing the hidden size of the fully-connected layers. 
It implies that the design of the mask branch is not specific, DCT-Mask is feasible for different setting of mask branch.

%\begin{table}[]
%\centering
%\begin{tabular}{c|c|c|c}
%\Xhline{2\arrayrulewidth}
%Conv & FC & FC size & AP   \\ \hline
%4    & 3  & 1024      & 36.5 \\
%4    & 4  & 1024      & 36.4 \\
%5    & 3  & 1024      & 36.3 \\
%4    & 3  & 2048      & 36.5 \\
%\Xhline{2\arrayrulewidth}
%\end{tabular}
%\vspace{0.1cm}
%\caption{The performance of different setting of mask branch including  the number of convolution layers, the number of fully-connected layers and the hidden size of fully-connected layers.}
%\label{table:mask branch}
%\end{table}

\begin{table*}[]
\centering
\scalebox{1.0}{
\begin{tabular}{l|c|c|c|l|c|c|l|l|l}
\Xhline{2\arrayrulewidth}
Method                  & Backbone        & aug. & sched.                   & AP   & AP@50 & AP@75 & AP$_S$  & AP$_M$  & AP$_L$  \\ \hline
MEInst\cite{zhang2020mask}                  & ResNet-101-FPN     & \checkmark    & 3$\times$                        & 33.9 & 56.2 & 35.4 & 19.8 & 36.1 & 42.3 \\ 
TensorMask\cite{chen2019tensormask}              & ResNet-101-FPN     & \checkmark    & 6$\times$   & 37.1 & 59.3 & 39.4 & 17.4 & 39.1 & 51.6 \\ 
MaskLab+\cite{chen2018masklab}                & ResNet-101-C4      & \checkmark    & 3$\times$                       & 37.3 & 59.8 & 39.6 & 16.9 & 39.9 & 53.5 \\ 
BMask R-CNN \cite{cheng2020boundary} & ResNet-101-FPN &     & 1$\times$ & 37.7 & 59.3 & 40.6 & 16.8 & 39.9 & 54.6 \\
MS R-CNN\cite{huang2019mask}        & ResNet-101-FPN  &      & 18e & 38.3 & 58.8 & 41.5 & 17.8 & 40.4 & 54.4 \\ 
BlendMask\cite{chen2020blendmask}               & ResNet-101-FPN     & \checkmark    & 3$\times$                        & 38.4 & 60.7 & 41.3 & 18.2 & 41.5 & 53.3 \\ 
Mask R-CNN\cite{he2017mask}              & ResNet-101-FPN  & \checkmark    & 3$\times$   & 38.8 & 60.9 & 41.9 & 21.8 & 41.4 & 50.5 \\ 
CondInst\cite{tian2020conditional}                & ResNet-101-FPN     & \checkmark    & 3$\times$   & 39.1 & 60.9 & 42.0   & 21.5 & 41.7 & 50.9 \\ 
SOLOv2\cite{wang2020solov2}                  & ResNet-101-FPN     & \checkmark    & 6$\times$                        & 39.7 & 60.7 & 42.9 & 17.3 & 42.9 & 57.4 \\ 
HTC\cite{chen2019hybrid}                     & ResNet-101-FPN  &      & 20e & 39.7 & 61.8 & 43.1 & 21.0   & 42.2 & 53.5 \\ 
HTC                     & ResNeXt-101-FPN &      & 20e & 41.2 & 63.9 & 44.7 & 22.8 & 43.9 & 54.6 \\
PointRender\cite{kirillov2020pointrend} & ResNeXt-101-FPN & \checkmark & 3$\times$ & 41.4 & 63.3 & 44.8 & 24.2 & 43.9 & 53.2 \\ \hline
%Mask R-CNN              & ResNet-50-FPN   & \checkmark    & 1$\times$   & 35.5 & 57   & 37.8 & 19.5 & 37.6 & 46   \\ \hline
%DCT-Mask R-CNN          & ResNet-50-FPN   & \checkmark    & 1$\times$                        & 36.7 & 56.6 & 39.8 & 20.3 & 38.9 & 47.1 \\ \hline
DCT-Mask R-CNN          & ResNet-101-FPN  & \checkmark    & 3$\times$                        & 40.1 & 61.2 & 43.6 & 22.7 & 42.7 & 51.8 \\ 
DCT-Mask R-CNN          & ResNeXt-101-FPN & \checkmark    & 3$\times$                        & 42.0   & 63.6 & 45.7 & 25.1 & 44.7 & 53.3 \\ 
Casecade DCT-Mask R-CNN & ResNet-101-FPN  & \checkmark    & 3$\times$                        & 41.0   & 61.7 & 44.7 & 23.7 & 43.3 & 52.6 \\ 
Casecade DCT-Mask R-CNN & ResNeXt-101-FPN & \checkmark    & 3$\times$                        & \textbf{42.6} & \textbf{64.0}   & \textbf{46.4} & \textbf{25.2} & \textbf{45.1} & 54.3 \\ \Xhline{2\arrayrulewidth}
\end{tabular}}
\vspace{-2mm}
\caption{Comparing different instance segmentation methods on COCO 2017 test-dev. ``aug.'': using multi-scale data augmentation during training. ``sched.'': the used learning rate schedule.}
\vspace{-2mm}
\label{table: sota}
\end{table*}

\begin{table}[]
\centering
\scalebox{1.0}{
\begin{tabular}{l|l|l|l}
\Xhline{2\arrayrulewidth}
Methods    & Backbone       & AP   & AP$_{50}$ \\ \hline
Mask R-CNN\cite{he2017mask} & ResNet-50-FPN &    26.2 & 49.9  \\ 
BshapeNet+\cite{kang2020bshapenet} & ResNet-50-FPN        & 27.3 & 50.5  \\ 
BMask R-CNN\cite{cheng2020boundary} & ResNet-50-FPN     & 29.4 & 54.7  \\ 
PointRend\cite{kirillov2020pointrend} & ResNet-50-FPN      & 30.4 & 55.1  \\ 
DCT-Mask    R-CNN & ResNet-50-FPN & 30.6 & 55.4  \\ \Xhline{2\arrayrulewidth}
\end{tabular}}
%\vspace{-2mm}
\caption{Comparing different instance segmentation methods on Cityscapes \emph{test} with only fine annotations.}
%\vspace{-4mm}
\label{table: cityscapes}
\end{table}

\section{More discussion to prior works}
In this section, we investigate relations and differences
between DCT-Mask and some prior works.

\textbf{Comparison with PointRend.}
PointRend\cite{kirillov2020pointrend} improves the mask representation by iteratively “rendering” the output mask from $7 \times 7$ resolution to $224\times 224$, while DCT-Mask directly achieves high resolution by decoding the DCT mask vector. As shown in Table \ref{table:PointRender}, our method achieves slightly better mask AP and outperforms PointRend by lower FLOPs and higher FPS. 

\textbf{Comparison with MEInst.}
%\begin{table}[!htbp]
%\centering
%\begin{tabular}{c|c|c|c|c}
%\Xhline{2\arrayrulewidth}
%Mask representation                & Resolution & Dim & AP & IoU \\ \hline
%Binary & $28\times 28$   &  784 &  30.8 & 0.938       \\ 
%PCA              & $28\times 28$  & 60 & 31.8 & 0.915       \\ 
%DCT              & $128\times 128$  & 300 & 32.3 & 0.970       \\ \Xhline{2\arrayrulewidth}
%\end{tabular}
%\caption{Comparison with MEInst on the COCO 2017val dataset. DCT outperforms PCA by using a higher resolution.}
%\label{table:MEInst}
%\end{table}
% PCA 128
MEInst\cite{zhang2020mask} proposes a one-stage instance segmentation framework based on FCOS, which applies PCA to encode the mask into a compact vector as mask representation. %We compare the performance of DCT with PCA in Table \ref{table:MEInst}. 
It increases the mask AP by reducing the training complexity with a much smaller compact vector. As we discussed before, mask quality is important to the mask prediction. MEInst only uses a $28\times 28$ resolution mask, and the mask quality drops to 91.5\%. We apply PCA to encode the mask with the same setting, which is $128\times 128$ resolution and 300 components. It only achieves 34.8 AP on COCO while DCT achieves 36.5 AP. Moreover, PCA needs extra pre-training. DCT is clearly a better choice for mask encoding.

%Besides, PCA is not suitable for high resolution mask. %For the $128\times 128$ resolution, the parameter amount of the covariance matrix in PCA is $128^4$ which is much more than the number of training samples. For $128\times 128$ resolution mask, PCA only achieves 82\% IoU  with a 300 dimensional vector, and 85\% with a 1000 dimensional vector. %DCT mask representation outperforms
%PCA by a much higher quality mask representation. Furthermore, this experiment can also be seen as an application of our method in one stage instance segmentation frameworks. 
%This experiment also indicates that our method could be integrated into one stage instance segmentation frameworks. 
%Here we do not present the results of PCA with $128 \times 128$ resolution. This experiment also indicates that our method could be integrated into one stage instance segmentation frameworks. 

\begin{table}[]
\centering
\scalebox{0.8}{
\begin{tabular}{c|c|c|c|c|c}
\Xhline{2\arrayrulewidth}
Method         & Resolution & COCO & LVIS$^*$  & FLOPs & FPS \\ \hline
%Mask R-CNN       & $28\times 28$     & 35.2 & 37.6 & 0.5B & 23 \\
DCT-Mask R-CNN   & $128\times 128$   & 36.5 & 39.7 & 0.5B & 22 \\
PointRend & $224\times 224$   & 36.3 & 39.7  & 0.9B & 16 \\ 
\Xhline{2\arrayrulewidth}
\end{tabular}}
%\vspace{-2mm}
\caption{Comparison with PointRend on the COCO 2017val dataset. DCT-Mask outperforms PointRend with lower FLOPs and higher FPS.}
%\vspace{-5mm}
\label{table:PointRender}
\end{table}
\section{Conclusion}
In this work, we have introduced a simple and effective approach to significantly increase the mask AP of instance segmentation. The increase is due to the high-quality and low-complexity DCT mask representation. 
DCT-Mask could be easily integrated into most pixel-based methods. We hope our method can serve as a fundamental approach in instance segmentation. 
% 最后说其他的压缩方法可能也会有用，dct 是用cos去拟合sigma函数，并不是完全合适的。也许会有更好的压缩方法。

\section{Acknowledgements}
This work was supported by Alibaba Innovative Research(AIR) Program and Major Scientific Research Project of Zhejiang Lab (No.2019DB0ZX01). 

Thank Xue Yang for his insightful help.

{\small
\bibliographystyle{ieee_fullname}
\bibliography{egbib}
}

\end{document}